\documentclass[10pt,twocolumn,letterpaper]{article}

\usepackage{iccv}
\usepackage{times}
\usepackage{epsfig}
\usepackage{graphicx}
\usepackage{amsmath}
\usepackage{amssymb}
\usepackage{multirow, adjustbox, booktabs, colortbl}
\usepackage{placeins}
\usepackage{subcaption}
\usepackage{sidecap}

\usepackage[pagebackref=true,breaklinks=true,letterpaper=true,colorlinks,bookmarks=false]{hyperref}

\iccvfinalcopy 

\ificcvfinal\fi

\begin{document}

\title{Semantic Diversity Learning for Zero-Shot Multi-label Classification}

\author{Avi Ben-Cohen \hspace{0.1cm} Nadav Zamir \hspace{0.1cm} Emanuel Ben Baruch \hspace{0.1cm} Itamar Friedman \hspace{0.1cm} Lihi Zelnik-Manor \\
DAMO Academy, Alibaba Group\\
\tt\small {\{avi.bencohen, nadav.zamir, emanuel.benbaruch, itamar.friedman, lihi.zelnik\}}\\
\tt\small{@alibaba-inc.com}
}

\maketitle
\ificcvfinal\thispagestyle{empty}\fi

\begin{abstract}

Training a neural network model for recognizing multiple labels associated with an image, including identifying unseen labels, is challenging, especially for images that portray numerous semantically diverse labels. As challenging as this task is, it is an essential task to tackle since it represents many real-world cases, such as image retrieval of natural images.
We argue that using a single embedding vector to represent an image, as commonly practiced, is not sufficient to rank both relevant seen and unseen labels accurately.
This study introduces an end-to-end model training for multi-label zero-shot learning that supports semantic diversity of the images and labels.
We propose to use an embedding matrix having principal embedding vectors trained using a tailored loss function.
In addition, during training, we suggest up-weighting in the loss function image samples presenting higher semantic diversity to encourage the diversity of the embedding matrix.
Extensive experiments show that our proposed method improves the zero-shot model’s quality in tag-based image retrieval achieving SoTA results on several common datasets (NUS-Wide, COCO, Open Images).

\end{abstract}

\section{Introduction}

\begin{figure}[t]
\centering
\includegraphics[width=\columnwidth]{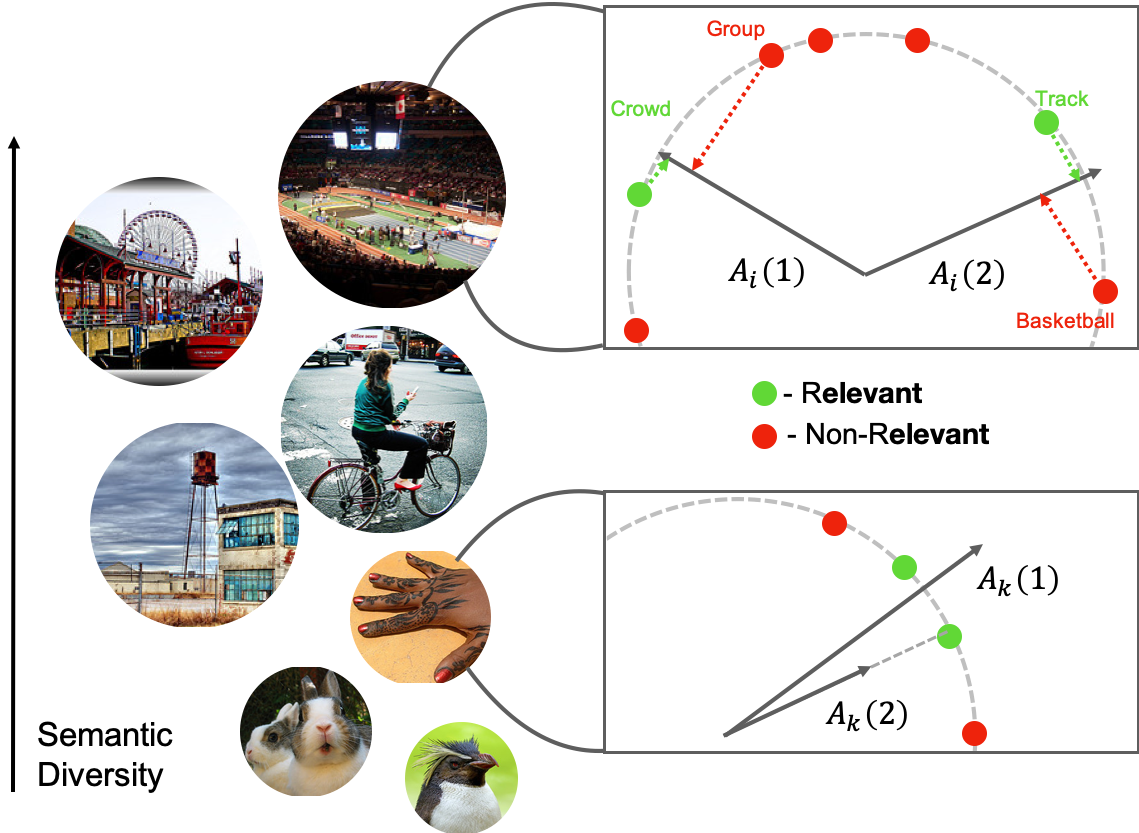}
\caption{Our model extracts a set of principal embedding vectors used as a transformation matrix $A_i$ where each row sets a ranking principal direction for labels in the word vector space based on their relevancy. By using multiple directions, it can deal with multiple diverse image semantic concepts. In addition, we propose that images with larger semantic diversity (top images) should be up-weighted during training compared to ones with lower semantic diversity (bottom images).
}
\label{fig:intro}
\end{figure}

Identifying all the relevant labels that describe the objects or scene in an image is an essential task in computer vision real-world applications. With the ongoing increase of photos stored online comes a growing need for better image tagging and tag-based retrieval for various use cases such as search, organization, or data collection. Recent datasets in this field enabled progress in this direction by introducing a large number of classes with annotations of their presence in each image. Yet, annotating a large number of classes for many images, each with high semantic diversity, can be very time-consuming and practically infeasible for real-world applications. Although current conventional multi-label classification methods can deal with a large number of classes, they are still limited by the annotated (seen) set of labels provided with the datasets.

On the other hand, Zero-shot (ZS) learning for multi-label classification adds the ability to recognize labels from additional categories that do not exist during training (unseen). This is usually done by transferring knowledge between the seen and unseen labels. In most cases, a text model \cite{bojanowski2016enriching, mikolov2013efficient, pennington2014glove} is used to transfer this knowledge using word vectors. Then, a visual model is trained, learning a transformation between the visual space and the word vector, label space.

Most studies on ZS classification focus on the single label problem, i.e., recognizing a single unseen label in each image \cite{akata2016multi, deutsch2017zero, frome2013devise, kodirov2017semantic, li2017zero, morgado2017semantically, xian2017zero, zhang2017learning}. However, a single label per image does not provide a full description of it and usually is not practical for real-world applications. Other studies tried to tackle the ZS multi-label classification problem. \cite{norouzi2013zero} trained a multi-label classifier on the seen classes and linearly combined word vectors of these classes using the prediction scores to represent an image. Based on that representative vector, the ranking of unseen labels was done by computing similarities to their word vectors. \cite{zhang2016fast} trained a network to output per-image a single principal direction that ranks relevant labels higher than non-relevant ones. However, this method faces difficulties with images that include multiple labels with high semantic diversity. In these cases, the extracted principal direction needs to be robust to high variability in the relevant labels' word vectors. For example, the classes \textit{"dog"} and \textit{"rice"} are quite different, but might still be present in the same image. As we grow with the number of annotated classes in images, the probable semantic diversity in each image grows. This high semantic diversity problem requires special treatment, which is hard to achieve using a single principal direction.

Several works had approached the problem of high semantic diversity of the labels in an image using a pre-trained object detector and learning to select bounding boxes of seen or unseen labels \cite{bansal2018zero, rahman2019deep0tag, rahman2019transductive, rahman2018zero}. Yet, these approaches require annotated bounding boxes as ground truth, making it not scalable for a large number of labels. Alternatively, \cite{huynh2020shared} used attention techniques to estimate the relevant regions based on a pre-trained model's features. However, this usually requires a large pre-trained model to get rich regional features (VGG-19) and a complex loss function to be tuned.

In this paper, we propose a method that aims to properly cope and leverage the semantic diversity of the labels in each image, by allowing multiple principal directions, constructed as a transformation matrix in the loss function. Also, sample images with larger semantic diversity are up-weighted in the loss function as these images are considered hard examples. As a result, our model learns to extract a per-image transformation designed to handle the image label diversity challenges (Figure \ref{fig:intro}). We believe that by doing so, we learn a model that is better suited for understanding and recognizing multiple seen and unseen labels in an image.

We further show how we achieve results that are on par or better than SoTA while keeping a relatively simple end-to-end training scheme using our suggested loss function.

The main contributions presented in this study include:
\begin{itemize}
\item A loss function tailored to the problem of ZS multi-label classification.
\item We show that up-weighting samples with higher semantic diversity further improves the model generalization.
\item An efficient end-to-end training scheme for ZS models is proposed, reaching SoTA results in tag-based image retrieval while keeping high-performance for image tagging on several datasets (NUS-Wide, Open-Images, and MS-COCO) with a smaller number of model parameters compared to other methods.
\end{itemize}

\section{Related Work}

Recent studies on multi-label classification reported notable success by exploiting dependencies among labels via graph neural networks to represent label relationships or word embeddings based on prior knowledge \cite{chen2019multi, chen2019multi2, durand2019learning,wang2020multi}. Other approaches try to model the image parts using attentional regions \cite{gao2020multi, wang2017multi, ye2020attention, you2020cross}. Although these approaches show promising results, they usually include a complex architecture, and other approaches reported similar and even better results using a more conventional training flow with advanced loss modifications \cite{ben2020asymmetric}. While most of these approaches are effective for images, including the seen classes they were trained for, they don't generalize well to unseen classes.

The main objective of zero-shot learning is to overcome this challenge and extract both seen and unseen labels for each image. This is usually done using semantic label information like attributes \cite{jayaraman2014zero} or word vector representation \cite{frome2013devise, akata2015label}. The central concept is to combine the visual features with the semantic word vectors representing each label using a similarity metric. Based on the similarity, unseen labels could be classified \cite{xian2017zero, romera2015embarrassingly,xian2016latent}. Most of the methods for zero-shot learning concentrate on finding the most dominant label in an image \cite{xian2018feature, xian2017zero, schonfeld2019generalized}. Despite their great success, these solutions do not generalize well to the problem of zero-shot multi-label classification and do not tackle the multi-label diversity challenges included in it.

In contrast to the zero-shot single-label classification task, multiple seen/unseen labels are assigned to an image in the zero-shot multi-label classification task. There is a limited number of studies addressing this problem. An interesting concept was suggested by \cite{norouzi2013zero} where predictions of a classifier trained on seen tags were linearly combined in the word embedding space to form a semantic embedding for that image to tackle the zero-shot single-label classification problem. This semantic embedding was later used to rank unseen labels based on their word vector's similarity to that embedding vector. \cite{li2015zero} extended their work by proposing a hierarchical semantic embedding to make the label embedding more representative for the multi-label task. \cite{fu2015transductive} proposed a transductive learning strategy to promote the regression model learned from seen classes to generalize well to unseen classes. In the Fast0Tag approach \cite{zhang2016fast}, the authors proposed a fast zero-shot tagging method by estimating a principal direction for an image. They show that word vectors of relevant tags in a given image rank ahead of the irrelevant tags along this principal direction in the word vector space. Another approach suggested by \cite{lee2018multi}, is using structured knowledge graphs to describe the relationships between multiple labels from the semantic label space and show how it can be applied to multi-label and zero-shot multi-label classification tasks. Due to the difficulty in distinguishing between multiple instances in an image using only global features, some studies try to identify important sub-regions in the image that includes the relevant labels by utilizing region proposal methods \cite{ren2017multiple, rahman2019deep0tag}. In recent work,  \cite{huynh2020shared} proposed a shared multi-attention model
for multi-label zero-shot learning that can focus on the relevant regions, obviating the need for object detection or region proposal networks. VGG-19 backbone is used to extract rich regional features, and a 4-term loss function is formed to tackle multiple challenges encountered during training. The derived model is then used to extract multiple attentions projected into the joint visual-label semantic embedding space to determine their labels. While this method tackles the diversity challenge by using multiple attention features for comparison to seen and unseen tags, it includes a complicated loss function consisting of a ranking loss and 3 regularization terms that require careful parameter tuning during training, while the image label diversity isn't used implicitly in any of these loss functions.

Finally, in our proposed method, we use the image semantic label diversity directly during training to improve and generalize our model better to diverse images. As far as we know, this is the first work to analyze the zero-shot semantic diversity problem and offer a method to exploit this information in a novel loss function. In addition, our end-to-end training flow does not require a large backbone model or object proposals for training while still achieving state-of-the-art results.

\section{Semantic Diversity Learning}
In this section, we present our proposed method for training multi-label zero-shot models. The problem and network architecture will be presented first, following a detailed description of our semantic diversity-based loss function.
\subsection{Problem Setting}
Let us denote by $\mathcal{S}$ and $\mathcal{U}$ the seen and unseen sets of tags, respectively, where seen tags stand for tags that have been seen during training and 'unseen' means tags that were not included in the training annotations. The entire set of tags is defined by $\mathcal{C}=\mathcal{S} \cup \mathcal{U}$.

Let $\{(I_n, Y_n); n =
1, 2,... , N\}$ denote the training data where $I_n$ is the $n$-th image, and $Y_n$ is the corresponding set of seen tags. We assume that each tag will be represented by a semantic word vector $\{v^c\}_{c \in \mathcal{C}}$. Based on these notations, we define the task of multi-label zero-shot learning as assigning the relevant unseen tags $y_i \subset{\mathcal{U}}$ for a given image $I_i$, and generalized multi-label zero-shot learning as assigning the relevant seen or unseen tags $y_i \subset{\mathcal{C}}$ for a given image $I_i$.

\subsection{Network Architecture}

\begin{figure*}[t]
\centering
\includegraphics[width=1.8\columnwidth]{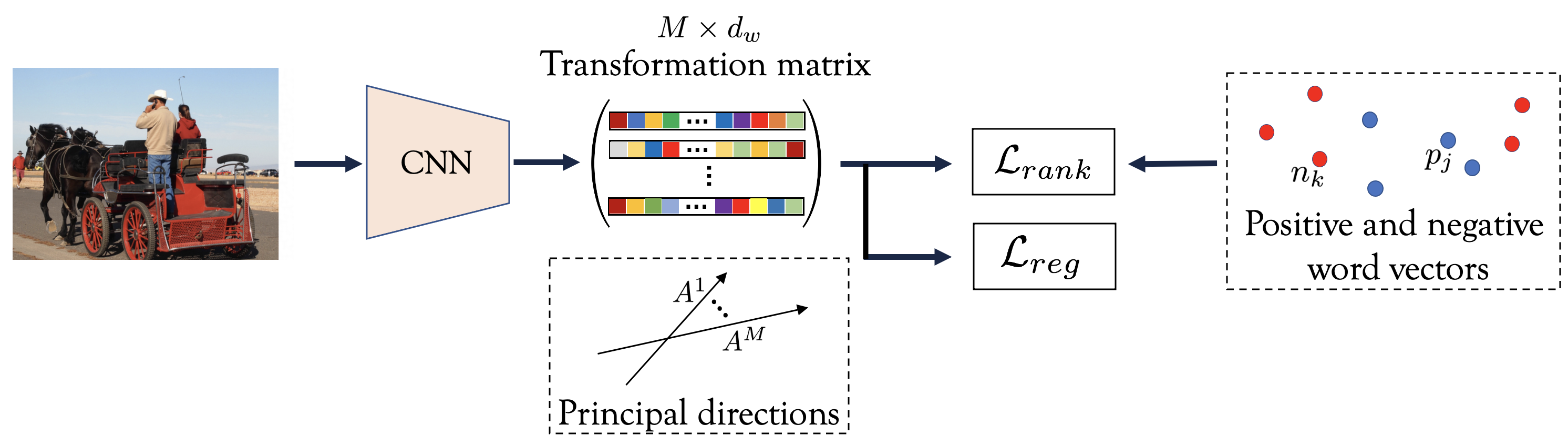}
\caption{The overview of our end-to-end training scheme for multi-label zero-shot learning. A CNN model is used to extract a per-image matrix transformation $A_i$ that includes several principal directions. $\mathcal{L}_{rank}$ loss pushes $A_i$ towards ranking positive labels higher than negative ones, and $\mathcal{L}_{reg}$ regularizes its principal directions.
}
\label{fig:framework}
\end{figure*}

The proposed network architecture is illustrated in Figure  \ref{fig:framework}. We used TResNet-M \cite{ridnik2021tresnet} convolutional neural network (CNN) as a backbone for our visual model. TResNet-M is a GPU-optimized model that reports high accuracy and efficiency on several competitive computer vision datasets. Using an efficient model design allows us to train easily in an end-to-end manner. By modifying its last fully connected layer, the vision model is trained to output a $M \times d_w$ linear transformation matrix where $d_w$ is the length of the word vectors, and $M$ is a parameter of our architecture representing the number of principal direction vectors. This matrix enables an image-dependent word ranking by projecting word vectors in different directions and using a ranking criteria over it. A similar concept was presented in \cite{yeh2019multilabel} for multi-label classification, where the transformation was learned per image and optimized to reach a linear combination of word vectors that allows it to distinguish the relevant from the non-relevant tags. However, in our experiments, simply using the method suggested in \cite{yeh2019multilabel} for the zero-shot learning task, the resulting model failed to generalize well for the unseen tags. Hence, we propose a loss function as well as ranking criteria more suitable for this task.

\subsection{Loss}
The per-image linear transformation $A$ extracted from the vision model should provide a high rank for all relevant tags even when there exists a large difference in their word embedding representation. Hence, we design our loss function to deal with the semantic diversity problem in zero-shot multi-label learning. Namely, we would like to provide a higher rank for a positive word vector $p_j$ and a lower rank for a negative word vector $n_k$, hence minimize the following:
\begin{equation}
\label{eq:v_diff}
u_{jk} = \max{(An_{k})} - \max{(Ap_{j})}
\end{equation}

The usage of a $\max$ function is crucial for this purpose as it allows each row in the matrix (principal direction) to be optimized in a different direction than other rows. In this formulation, one matrix row is sufficient to rank a label correctly, thus letting other rows focus on the additional relevant labels and output high scores for all of them.
This is ideal if there is a high semantic diversity in the image that makes it difficult for a single row in the transformation matrix to deal with the multiple and diverse set of tags. Similar intuition is presented in multi-class support vector machines (SVM) \cite{crammer2001algorithmic}, where the confidence value for the correct label is optimized to be larger by a certain margin than the confidences assigned to the rest
of the labels. Otherwise, we get a loss that is linearly proportional to the difference between the confidence of the correct label and the maximum among the other labels' confidences. Also, by using the $\max$ operation when comparing a specific pair of positive and negative labels, we allow gradients propagation only through the most dominant row in the matrix for each tag, hence, not modifying the entire matrix for each pair which allows each row to focus on different semantic concepts. 

Having this in mind, we define the ranking loss, inspired by \cite{zhang2016fast}, as follows:

\begin{equation}
\label{eq:L_rank}
\mathcal{L}_{rank} = \omega_{d}\frac{1}{\omega_n}\sum_{j}\sum_{k}\log{(1+e^{u_{jk}})}
\end{equation}
Where $\omega_n = |P||\bar{P}|$, while $|P|$ is the size of the set of ground-truth tags for a single image and $|\bar{P}|$ is the size of the set of the non relevant tags. The division by $\omega_n$ is used as a normalization. Images with a large number of tags will be treated the same as these with a low number of tags. We consider images with high label diversity more difficult as they require our model to learn how to rank several different semantic concepts higher than others. Hence, we would like to put additional focus on these examples during training. We denote by $\omega_{d}$ the per-image semantic diversity weight (SDW). The SDW up-weights more diverse images (hard samples), thus increasing focus on them in our loss function. We define $\omega_{d}$ as the sum of variances across the relevant set of tags in an image:
\begin{equation}
\label{eq:omega_d}
\omega_{d} = 1+\sum_{i=1}^{d_w}{\text{var}(P_i)}
\end{equation}

We note that our ranking loss function provides high flexibility when learning the transformation matrix which is good for learning diverse tags per image. However, this flexibility should be controlled when training on large and noisy datasets. The matrix rows could become too diverse and, by such, making it sensitive to outliers and drive it to not generalize well when training on large and noisy datasets. Hence, we add a regularization term to deal with these cases defined by
\begin{equation}
\label{eq:L_reg}
\mathcal{L}_{reg} =  {\Bigg\lVert \sum_{m=1}^{d_w}{\text{var}(A^m)}\Bigg\rVert}_{1} 
\end{equation}

This regularization term puts a constraint on the matrix rows from being too diverse. In practice, reducing the variance between rows encourages learning correlative information between tags known to be useful in multi-label setting \cite{chen2019multi2}. This regularization term is similar in a way to multi-class SVM regularization \cite{crammer2001algorithmic}. However, by looking at the variance, we ignore the mean of each column in the matrix, making it invariant to translations. This is different from the standard $l_2$ norm regularization used in multi-class SVM. The main reasoning is that our loss function is optimized over a given word vector space that its features are not standardized. Hence, using the variance as a regularization instead of the standard $l_2$ norm is more suitable for our use case. Our final loss function is defined by
\begin{equation}
\label{eq:L_final}
\begin{split}
\mathcal{L}_{final} =  \frac{1}{N}\sum_{i=1}^{N}\Big[(1-\tilde{\lambda}){\mathcal{L}_{rank}(A_i,y_i)} +  \tilde{\lambda}{\mathcal{L}_{reg}(A_i)}\Big]
\end{split}
\end{equation}
Where $\tilde{\lambda}$ sets the regularization weight. In practice we use $\lambda = \tilde{\lambda}|\bar{P}|$ as the regularization parameter which is invariant to the number of negative tags per image.

\begin{table*}[t]
\centering
\caption{State-of-the-art comparison for ZSL and GZSL tasks on the NUS-WIDE and Open Images datasets. We report the results in terms of mAP, as well as precision (P), recall (R), and F1 score at $K{\in}\{3,5\}$ for NUS-WIDE and $K{\in}\{10,20\}$ for Open Images. Best results are in bold. }
\adjustbox{width=\linewidth}{
\begin{tabular}{ccccccccc|ccccccc} 
\toprule[0.15em]
\multirow{3}{*}{\textbf{ Method}} & \multirow{3}{*}{\textbf{Task }} & \multicolumn{7}{c}{\textbf{NUS-WIDE ( \#seen / \#unseen = 925/81) }} & \multicolumn{7}{c}{\textbf{Open-Images ( \#seen / \#unseen = 7186/400) }} \\
 &  & \multicolumn{3}{c}{\textbf{K = 3 }} & \multicolumn{3}{c}{\textbf{K = 5 }} & \multirow{2}{*}{\textbf{mAP }} & \multicolumn{3}{c}{\textbf{K = 10 }} & \multicolumn{3}{c}{\textbf{K = 20 }} & \multirow{2}{*}{\textbf{mAP }} \\
 &  & \textbf{P } & \textbf{R } & \textbf{F1 } & \textbf{P } & \textbf{R } & \textbf{F1 } &  & \textbf{P } & \textbf{R } & \textbf{F1 } & \textbf{P } & \textbf{R } & \textbf{F1 } &  \\ 
\toprule[0.15em]
\multirow{2}{*}{CONSE~\cite{norouzi2013zero}} & ZSL & 17.5 & 28.0 & 21.6 & 13.9 & 37.0 & 20.2 & 9.4 & 0.2 & 7.3 & 0.4 & 0.2 & 11.3 & 0.3 & 40.4 \\
 & GZSL & 11.5 & 5.1 & 7.0 & 9.6 & 7.1 & 8.1 & 2.1 & 2.4 & 2.8 & 2.6 & 1.7 & 3.9 & 2.4 & 43.5 \\ 
\cmidrule(lr){2-16}
\multirow{2}{*}{LabelEM~\cite{akata2015label}} & ZSL & 15.6 & 25.0 & 19.2 & 13.4 & 35.7 & 19.5 & 7.1 & 0.2 & 8.7 & 0.5 & 0.2 & 15.8 & 0.4 & 40.5 \\
 & GZSL & 15.5 & 6.8 & 9.5 & 13.4 & 9.8 & 11.3 & 2.2 & 4.8 & 5.6 & 5.2 & 3.7 & 8.5 & 5.1 & 45.2 \\ 
\cmidrule(lr){2-16}
\multirow{2}{*}{Fast0Tag~\cite{zhang2016fast}} & ZSL & 22.6 & 36.2 & 27.8 & 18.2 & 48.4 & 26.4 & 15.1 & 0.3 & 12.6 & 0.7 & 0.3 & 21.3 & 0.6 & 41.2 \\
 & GZSL & 18.8 & 8.3 & 11.5 & 15.9 & 11.7 & 13.5 & 3.7 & 14.8 & 17.3 & 16.0 & 9.3 & 21.5 & 12.9 & 45.2 \\ 
\cmidrule(lr){2-16}
\multirow{2}{*}{One Attention per Label~\cite{kim2018bilinear}} & ZS & 20.9 & 33.5 & 25.8 & 16.2 & 43.2 & 23.6 & 10.4 & - & - & - & - & - & - & - \\
 & GZSL & 17.9 & 7.9 & 10.9 & 15.6 & 11.5 & 13.2 & 3.7 & - & - & - & - & - & - & - \\ 
\cmidrule(lr){2-16}
\multirow{2}{*}{LESA (M=10)~\cite{huynh2020shared}} & ZSL & \textbf{25.7} & {41.1} & \textbf{31.6} & \textbf{19.7} & {52.5} & \textbf{28.7} & 19.4 & 0.7 & 25.6 & 1.4 & 0.5 & 37.4 & 1.0 & 41.7 \\
 & GZSL & 23.6 & 10.4 & 14.4 & 19.8 & 14.6 & 16.8 & 5.6 & 16.2 & 18.9 & 17.4 & 10.2 & 23.9 & 14.3 & 45.4 \\ 
 \cmidrule(lr){2-16}
  
 \multirow{2}{*}{\textbf{Ours} (M=7)} & ZSL & {24.2}  & {\textbf{41.3}}  & {30.5}  & {18.8}  & {\textbf{53.4}}  & {27.8}  & {\textbf{25.9}}  & \textbf{{6.1}} & \textbf{{47.0}} & \textbf{{10.7}} & \textbf{{4.4}}  & \textbf{{68.1}} & \textbf{{8.3}} & \textbf{{62.9}} \\
 & GZSL & {\textbf{27.7}}  & {\textbf{13.9}}  & {\textbf{18.5}}  & {\textbf{23.0}}  & {\textbf{19.3}}  & {\textbf{21.0}}  & {\textbf{12.1}}  & \textbf{{35.3}} & \textbf{{40.8}} & {\textbf{37.8}} & \textbf{{23.6}} & {\textbf{54.5}} & {\textbf{32.9}} & {\textbf{75.3}} \\
  \cmidrule(lr){2-16}
 
\bottomrule[0.1em]
\end{tabular}
}
\vspace{-0.15cm}
\label{tab:sota_nuswide_openimages}
\end{table*}

\section{Experiments}
Several experiments were conducted to analyze and evaluate our method for zero-shot multi-label classification. In section \ref{sec:sota} we compare our approach to other state-of-the-art works in the field. To better understand each component's contribution in our framework, we conduct an ablation study as discussed in section \ref{sec:ablation}, and the regularization parameter is further analyzed in section \ref{sec:regularization}. Next, to visualize what the transformation matrix learns using our method, a set of qualitative results are presented and discussed in section \ref{sec:qualitative}. As our method aims towards dealing with the semantic diversity challenge, we wish to analyze our results on the more diverse set of images as discussed in section \ref{sec:diverse}. Finally, in section \ref{sec:rows} we discuss and analyze our results using a different number of principal directions in the transformation matrix.

\begin{table}[t]
\centering
\caption{State-of-the-art comparison on the MS COCO dataset split into $48$ seen and $17$ unseen classes. We report the results in terms of precision (P), recall (R), and F1 score at $K{=}3$ for ZSL and GZSL tasks.}\vspace{0.2em} 
\setlength{\tabcolsep}{12pt}
\adjustbox{width=1\linewidth}{
\begin{tabular}{ccccc} 
\toprule[0.15em]
\textbf{Method} & \textbf{Task} & \begin{tabular}[c]{@{}c@{}} \textbf{P} \end{tabular} & \begin{tabular}[c]{@{}c@{}} \textbf{R} \end{tabular} & \textbf{F1} \\
\toprule[0.15em]
\multirow{2}{*}{CONSE~\cite{norouzi2013zero}} & ZSL & 11.4 & 28.3 & 16.2 \\
 & GZSL & 23.8 & 28.8 & 26.1 \\ 
\cmidrule(r){2-5}
\multirow{2}{*}{Fast0tag~\cite{zhang2016fast}} & ZSL & 24.7 & 61.4 & 25.3 \\
 & GZSL & 38.5 & 46.5 & 42.1 \\ 
\cmidrule(r){2-5}
\multirow{2}{*}{Deep0tag~\cite{rahman2019deep0tag}} & ZSL & 26.5 & 65.9 & 37.8 \\
 & GZSL & 43.2 & 52.2 & 47.3 \\ 
\cmidrule(r){2-5}
\multirow{2}{*}{\textbf{Ours} (M=2)} & ZSL & {26.3} & 65.3 & {37.5} \\
 & GZSL & {59.0} & {60.8} & {59.9} \\
 
\bottomrule[0.1em]
\end{tabular}%
}
\vspace{-0.2cm}
\label{tab:sota_coco}
\end{table}

\begin{table*}[t]
	\centering
	\setlength{\tabcolsep}{4pt}
	\begin{tabular}{r|ccc|ccccc|c}
    	 &     \multicolumn{3}{c|}{Fast0Tag \cite{zhang2016fast}}\\
    &Orig&Impl. & Base. & a & b  & c & d & f   & Ours \\
    	 \hline
        SDW & & & & \checkmark & \checkmark & \checkmark & \checkmark & &\checkmark   \\
        M=2  & & & &  & \checkmark & \checkmark &  &  & \\
        Reg. (0.1) &  &   &  &  &  & \checkmark & \checkmark & &   \\
        M=7  &  & & & & &  & \checkmark &\checkmark&\checkmark   \\
        Reg. (0.3)              & & &  &  &  &  &  &  \checkmark&\checkmark   \\
        \hline
        \multirow{ 2}{*}{mAP} GZS
                         & 3.7 & 9.7 & 9.5 & 10.2 & 10.6 & 11.0 & 12.2 & 11.8 & 12.1    \\
        ZS              & 15.1 & 21.6 & 20.6 & 22.7 & 22.2 & 23.8 & 25.1 & 25.8 & 25.9    \\
    
	\end{tabular}
    \medskip
    \caption{Ablation study showing the contribution of the different components in our training scheme compared to the loss presented in Fast0Tag showing the original implementation results, our implementation results using our training framework, and the baseline on NUS-Wide test set.}
    \label{tab:ablation}
\end{table*}

\subsection{Setup}
\noindent\textbf{Datasets:} Three datasets were used to evaluate our proposed methodology.
The \textbf{NUS-WIDE} \cite{nuswide} dataset includes $270$K images with $81$ human-annotated categories used as unseen classes in addition to a set of $925$ labels obtained from Flickr users tags automatically that are used as seen classes.
The \textbf{MS COCO} \cite{coco} dataset is divided into training and validation sets with $82{,}783$ and $40{,}504$ images, respectively. This dataset is commonly used for multi-label zero-shot object detection~\cite{bansal2018zero,hayat2020synthesizing} and was also used in recent works of multi-label zero-shot classification \cite{rahman2019deep0tag}. We follow \cite{bansal2018zero} with our split to seen and unseen tags, resulting in $48$ seen and $17$ unseen classes based on their cluster embedding in the semantic space and WordNet hierarchy \cite{miller1995wordnet}. We use the provided list of images, including $73{,}774$ images with only seen objects for training, and $6{,}608$ images containing both seen and unseen objects for testing.

The \textbf{Open Images} (v4) \cite{openimages} dataset consists of $9$ million training images, $41{,}620$  validation images, and $125{,}456$ test images. This dataset introduces several challenges: this large-scale dataset is larger by orders of magnitude when compared to NUS-WIDE or MS COCO, and its images are only partially annotated where not all labels were verified as true-positives or negatives in each image. Similar to \cite{huynh2020shared}, we use $7{,}186$ labels, having at least $100$ images in training set for each seen class. The most frequent $400$ test labels not present in the training data are selected as unseen classes. 

\noindent\textbf{Evaluation Metrics:} We follow \cite{huynh2020shared} and use the mean Average Precision (mAP) and F1 score at \textit{top}-$K$ predictions in each image. The mAP evaluates the accuracy for tag-based retrieval, i.e., it answers the question of how good our model is at ranking images for each given label, while the \textit{top}-$K$ F1 score captures its accuracy for image tagging, measuring how good it is at ranking relevant labels for each image. 

\noindent\textbf{Implementation Details:}
Unless stated otherwise,  all experiments were conducted with the following training configuration. We use as a backbone TResNet-M, pre-trained on the ImageNet dataset \cite{imagenet_cvpr09}. See appendix \ref{sec:more_experiments} for a comparison to other backbones. The model was fine-tuned using Adam optimizer \cite{kingma2017adam} and 1-cycle cosine annealing policy \cite{smith2018disciplined} with maximal learning rate of 1e-4. We use cutout \cite{devries2017improved} with probability of 0.5, True-weight-decay \cite{loshchilov2019decoupled} of 3e-4 and standard ImageNet augmentations. The regularization parameter $\lambda$ was set to $0.3$. We train the network for 7/7/20 epochs and a batch-size of 192/96/32 for NUS-Wide/Open Images/MS-COCO, respectively.

For our tag embedding representations we use a FastText pre-trained model \cite{grave2018learning} with a vector size of $d_w=300$. The word vectors are $\ell_2$ normalized. At inference, our trained model takes an image $I_i$ as input and provides a corresponding transformation matrix $A_i$ as output. Let $T={\{t_j\}}$ denote the set of word vectors representing each tag in the label set. For image tagging, we compute $r_{ij}=\max(A_{i}t_{j})$ for each seen/unseen tag and rank them such that higher values represent the more relevant tags. For tag-based image retrieval, for a query tag, we compute similarly $r_{ij}$ for all given images and rank them from most relevant to least. 

\noindent\textbf{Baseline:} We use as a baseline to our method implementation of Fast0Tag \cite{zhang2016fast} loss function integrated within our training framework. In addition, we also compare our method to a baseline with multiple principal directions ($M=7$), substituting our $\max$ function in equation \eqref{eq:v_diff} with $l_2$ norm similar to \cite{yeh2019multilabel}, and removing our regularization term and SDW.

\subsection{Comparison to State-of-the-art}
\label{sec:sota}
Table \ref{tab:sota_nuswide_openimages} shows a comparison of our proposed method to other state-of-the-art methods on NUS-WIDE and Open Images. Our method outperforms all other methods in terms of mAP for both datasets. We also present the \textit{top}-$K$ Precision (P) and Recall (R) in addition to the F1 score. Note that we used $K\in{\{3,5\}}$ for NUS-WIDE and $K\in{\{10,20\}}$ for Open Images due to a large number of available labels in it. Compared to the recently introduced shared multi-attention-based approach (LESA) \cite{huynh2020shared} we achieve better performance on open images. We improved results in both zero-shot/generalized zero-shot learning tasks by 9.3\%/20.4\%, 7.3\%/18.6\%, and 21.2\%/29.9\% in $F1(K=10)$, $F1(K=20)$, and $mAP$ respectively.

As for NUS-WIDE, although the LESA approach shows a moderate improvement in performance in terms of F1 for zero-shot learning, our proposed method shows a much higher gain in terms of mAP with an improvement of 6.5\% for both zero-shot and generalized zero-shot. In addition, our method achieves improved results in terms of F1 for generalized zero-shot of 4.1\% and 4.2\% for $F1(K=3)$, $F1(K=5)$ respectively.

During this paper's writing, recent work in the field was published as a pre-print showing a Generative approach for zero-shot learning \cite{gupta2021generative}. In this approach, the authors propose to train two separate classifiers, one focused on the zero-shot learning task and another on the generalized zero-shot learning task. This differs from our problem formulation, as we wish to have a single model that is trained for both zero-shot and generalized zero-shot tasks. The formulation used in our study seems to be more suitable for real-world applications, i.e., evaluating one single model for the two tasks under the same working point.

MS-COCO dataset is quite different from NUS-Wide and Open Images as it holds a relatively small number of seen and unseen labels. In Table \ref{tab:sota_coco} we compare to the method presented in \cite{rahman2019deep0tag} which is based on an object detection model. In comparison, our model achieves slightly lower results for zero-shot learning with significant improvement in the generalized zero-shot metrics. As presented in Figure \ref{fig:framework}, our framework does not include any additional modules such as object detectors, region proposals, or attention layers and can still achieve high-quality results. Note that for COCO, we used $M=2$. We discuss the motivation for this in section \ref{sec:rows}.

\begin{figure*}
\centering
\begin{tabular}{lll}
\subcaptionbox*{}{\includegraphics[height = 2.6cm]{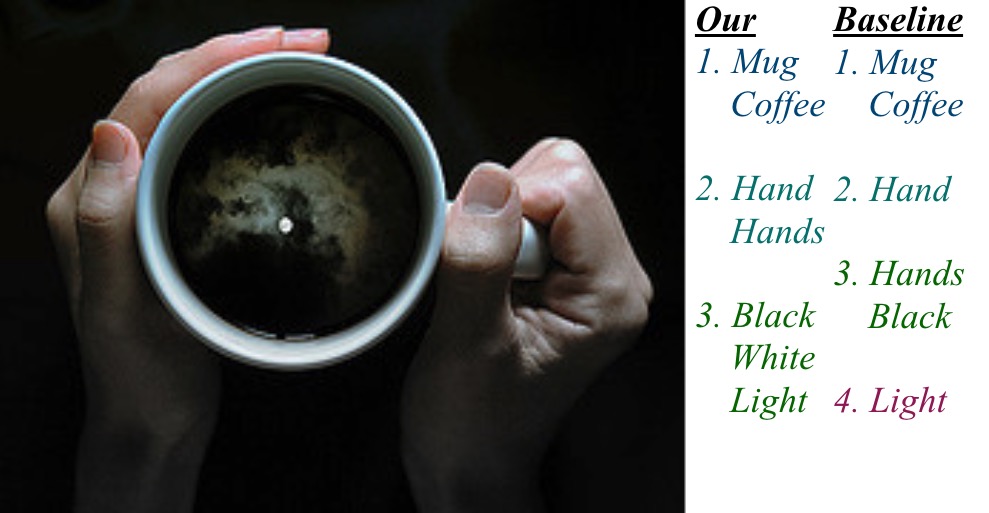}} &
\subcaptionbox*{}{\includegraphics[height = 2.6cm]{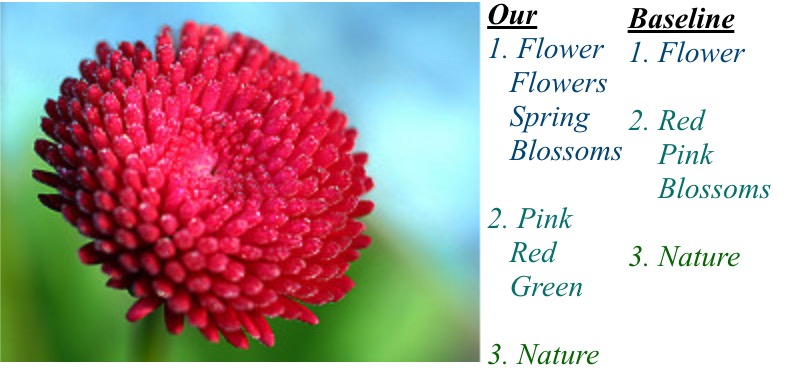}} &
\subcaptionbox*{}{\includegraphics[height = 2.6cm]{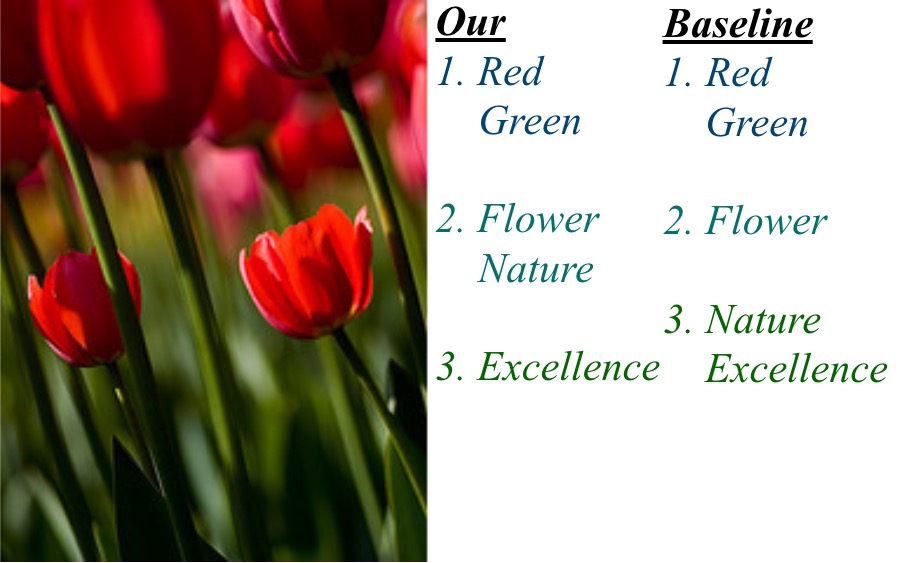}} \\
\subcaptionbox*{}{\includegraphics[height = 2.6cm]{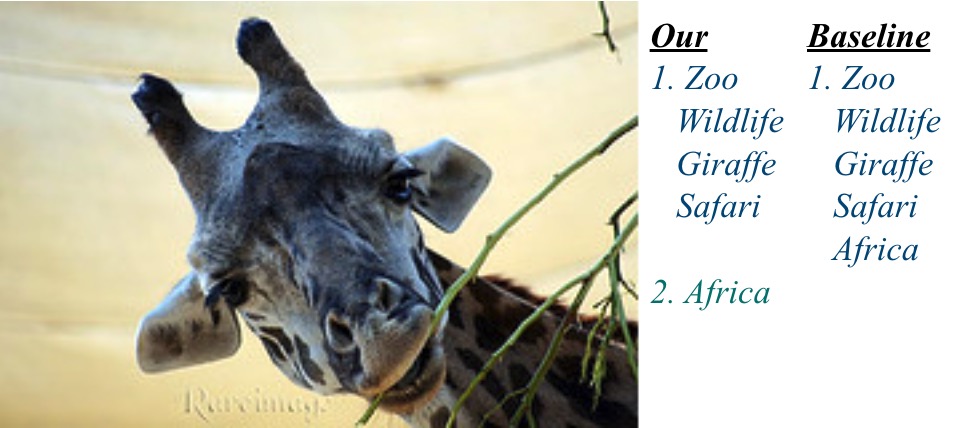}} &

\subcaptionbox*{}{\includegraphics[height = 2.6cm]{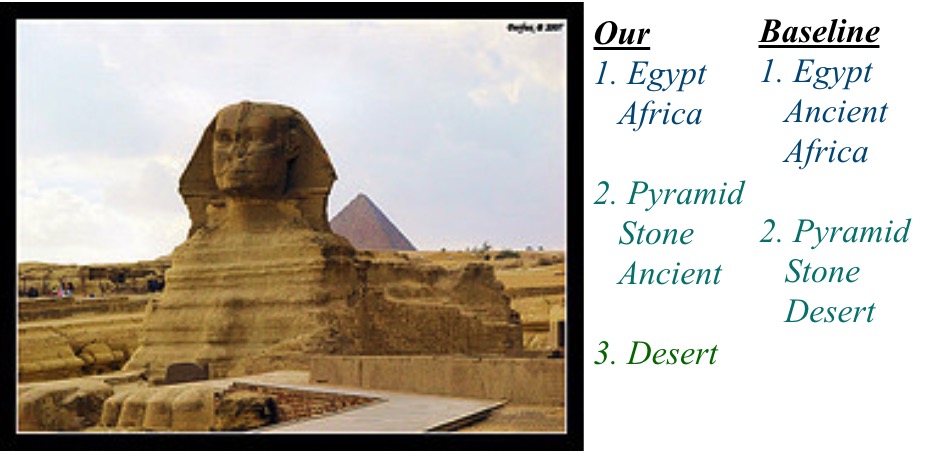}} &

\subcaptionbox*{}{\includegraphics[height = 2.6cm]{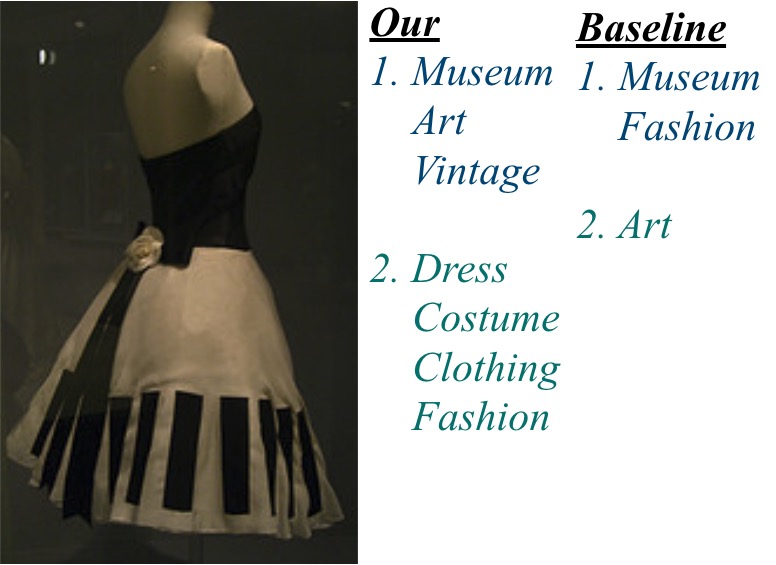}}

\end{tabular}
\caption{Qualitative results, showing the most relevant tags for each row in the transformation matrix using our proposed method and the baseline. Using our proposed method, we can see that different semantic concepts are being learned by different rows in the matrix.}
\label{fig:qualitative}
\end{figure*}

\begin{figure}
\centering
\begin{subfigure}[b]{0.4\textwidth}
  \includegraphics[width=1\linewidth]{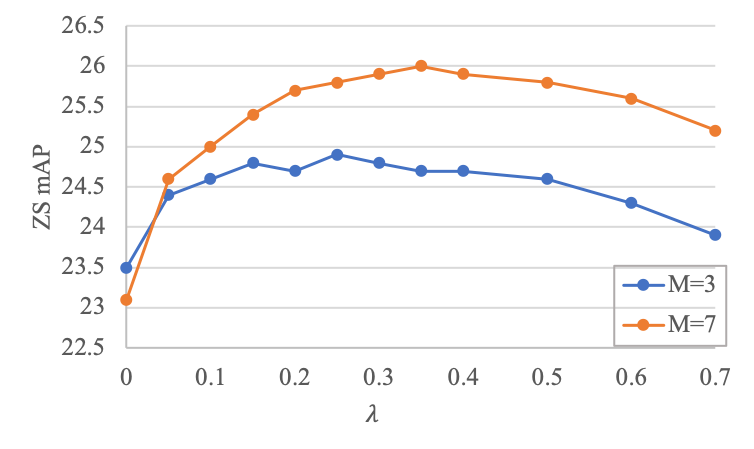}
  \caption{}
  \label{fig:reg_map} 
\end{subfigure}

\begin{subfigure}[b]{0.4\textwidth}
  \includegraphics[width=1\linewidth]{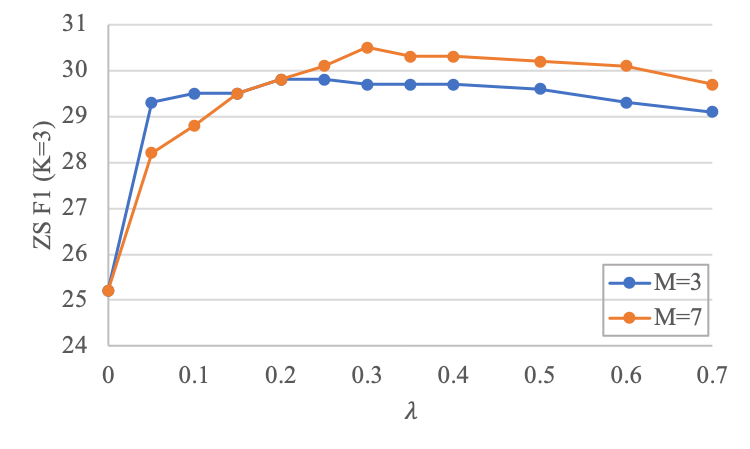}
  \caption{}
  \label{fig:reg_f1}
\end{subfigure}

\caption{Zero-shot multi label classification quality as a function of $\lambda$  and $M=\{3,7\}$ in the transformation matrix. (a) The mAP for the multi-label ZS task on NUS-wide test set; (b) Similar to (a) showing $F1 (K=3)$.}
\label{fig:regularization}
\end{figure}

\subsection{Ablation Study}
\label{sec:ablation}
To better understand each component's contribution in our solution, we perform an ablation study as shown in Table \ref{tab:ablation}. We compare to Fast0Tag \cite{zhang2016fast} method that uses a single principal direction per image as a starting point. We also implement Fast0Tag in our framework for a fair comparison, which already shows much better results compared to the original paper. However, the baseline with additional principal directions shows lower performance in terms of mAP, indicating that a naive addition of principal directions is insufficient to improve the model performance. The addition of SDW has been shown to improve the results when used together with the loss proposed in \cite{zhang2016fast}. It also improves the results of our proposed method (column f to ours), indicating that SDW can be beneficial for different methods and loss functions that can support per-sample weighting. In addition, the regularization term has shown to improve the results when using $M=2$ while for $M=7$ higher regularization showed higher performance in terms of ZS mAP while keeping on par results in terms of GZS mAP.

\subsection{Regularization Parameter}
\label{sec:regularization}
The regularization term presented in equation \eqref{eq:L_reg} provides control over the matrix transformation rows' diversity. Figure \ref{fig:regularization} presents the results in terms of mAP and F1 ($K=3$) on NUS-Wide test set using different $\lambda$ values for $M=\{3,7\}$. The contribution of the regularization term is noticeable for a different number of rows. However, for a larger number of rows ($M=7$) stronger regularization provides better performance.
A possible reason for that is that a large number of rows in the matrix can lead to a decrease in utilization of all rows in practice. Using our proposed regularization, we better utilize the different rows in the matrix and thus better generalize on the test set. Especially for zero-shot learning, the generalizability of the model is crucial for retrieving images with unseen tags.

\subsection{Qualitative Assessment}
\label{sec:qualitative}
Our proposed method focuses on semantic diversity learning using a matrix transformation. Each row in this matrix can be described as a principal direction responsible for a set of relevant labels. In Figure \ref{fig:qualitative} we compare our method results to the baseline model. For each image, we show the most relevant results in the $top-10$ retrieved labels. The numbers in the figure indicate the most dominant row that provided the highest score for the corresponding set of tags. In several of these sample images, we can see that the tags learned using our approach were separated based on their main concept, e.g., in the top left image, hand and hands belong to the same row, while for the baseline, they are separated. Moreover, using our method, we can see that more relevant tags were discovered in some cases, e.g., in the bottom right image, additional tags such as "clothing" and "dress" were discovered by the same row that learned to understand this concept in the image.

\subsection{Performance on Diverse Images}
\label{sec:diverse}
As our method aims towards learning diverse concepts in an image, we wish to evaluate its performance on the more-diverse samples in the dataset. Since diverse images usually include more labels, we perform an experiment to evaluate our image tagging method on image samples that include more than $6$ labels from both zero-shot and generalized zero-shot sets. Table \ref{tab:diverse} presents the results using the baseline and our proposed method ($M=7$). Since SDW up-weights more diverse images in the loss function, we show results both with and without it compared to the baseline. Our method outperforms the baseline without SDW and achieves even higher results when adding it during training, demonstrating its effectiveness with managing diverse samples.

\begin{table}[t]
\centering
\caption{ZS multi-label classification results with $M=7$ rows in the transformation matrix, for samples with more than 6 unseen labels, in terms of precision (P), recall (R), and F1 for top 10 retrieved labels on NUS-Wide test set.}\vspace{0.2em} 
\setlength{\tabcolsep}{15pt}
\adjustbox{width=1\linewidth}{
\begin{tabular}{cccc} 
\toprule[0.15em]
\textbf{Method}  & \begin{tabular}[c]{@{}c@{}} \textbf{P} \end{tabular} & \begin{tabular}[c]{@{}c@{}} \textbf{R} \end{tabular} & \textbf{F1} \\
\toprule[0.15em]
{Baseline}  & 31.7 & 44.8 & 37.1 \\
{Our w/o SDW}  & 36.2 & 51.2 & 42.4 \\
{\textbf{Our}}  & \textbf{36.6} & \textbf{51.9} & \textbf{42.9} \\

\bottomrule[0.1em]
\end{tabular}%
}
\vspace{-0.2cm}
\label{tab:diverse}
\end{table}

\subsection{Matrix Principal directions}
\label{sec:rows}
\begin{figure}[t]
\centering
\includegraphics[width=\columnwidth]{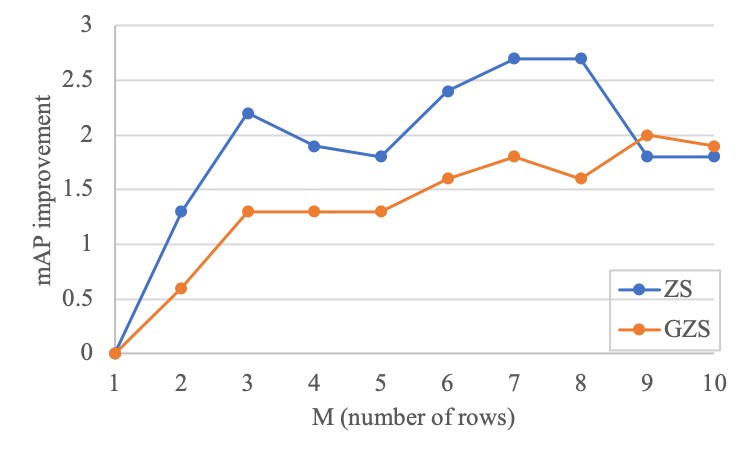}
\caption{mAP improvement with different number of rows ($M$) in the transformation matrix compared to $M=1$ for zero-shot and generalized zero-shot tasks on NUS-Wide test set.
}
\label{fig:mAPvsRow}
\end{figure}

The size of the per-image transformation matrix $A$ is set by the parameter $M$ that defines the number of principal directions it has. In Figure \ref{fig:mAPvsRow} the mAP improvement in terms of zero-shot and generalized zero-shot is presented for a different number of rows in the matrix using a fixed set of parameters (e.g., regularization parameter = $0.1$) on the NUS-Wide test set. 

Noticeably, an increasing number of rows improves the generalized zero-shot results. While the set of principal directions used increases, it is easier for the model to learn the seen tags in the image and improve the generalized zero-shot performance. As for zero-shot learning, while there is an improvement when increasing $M$ up to $7-8$ rows, a further increase in the number of rows causes a decrease in mAP. This indicates that the model does not generalize well to unseen tags in our method when using too many principal directions. A possible solution for this would be to increase the regularization parameter. In our experiments, we have found that $M=7$ covers the semantic diversity in the image compared to other choices for NUS-Wide and Open Images. While for MS-COCO, which is much smaller in the number of labels, $M=2$ was found experimentally to show superior results. 

\section{Conclusions}
The zero-shot multi-label classification task introduces the challenge of recognizing multiple and diverse labels in an image, including categories not seen during the training process. This task is even more challenging for images with high semantic diversity.

In this study, we propose an end-to-end training scheme including a novel loss function tailored to semantic diversity in zero-shot learning. Our loss function consists of a semantic diversity weight for more diverse images, utilizing multiple principal directions to enable specialization of principal vectors in different semantic concepts and a matrix variance regularization term to improve model generalizability to unseen tags. Extensive experiments show that our proposed loss function improves the zero-shot model's quality in terms of tag-based image retrieval showing SoTA results while keeping high-performance for image tagging on several standard datasets (NUS-Wide, COCO, Open Images).



\FloatBarrier
{\small
\bibliographystyle{ieee_fullname}
\bibliography{main}

\begin{thebibliography}{10}\itemsep=-1pt

\bibitem{akata2016multi}
Zeynep Akata, Mateusz Malinowski, Mario Fritz, and Bernt Schiele.
\newblock Multi-cue zero-shot learning with strong supervision.
\newblock In {\em Proceedings of the IEEE Conference on Computer Vision and
  Pattern Recognition}, pages 59--68, 2016.

\bibitem{akata2015label}
Zeynep Akata, Florent Perronnin, Zaid Harchaoui, and Cordelia Schmid.
\newblock Label-embedding for image classification.
\newblock {\em IEEE transactions on pattern analysis and machine intelligence},
  38(7):1425--1438, 2015.

\bibitem{bansal2018zero}
Ankan Bansal, Karan Sikka, Gaurav Sharma, Rama Chellappa, and Ajay Divakaran.
\newblock Zero-shot object detection.
\newblock In {\em Proceedings of the European Conference on Computer Vision
  (ECCV)}, pages 384--400, 2018.

\bibitem{ben2020asymmetric}
Emanuel Ben-Baruch, Tal Ridnik, Nadav Zamir, Asaf Noy, Itamar Friedman, Matan
  Protter, and Lihi Zelnik-Manor.
\newblock Asymmetric loss for multi-label classification.
\newblock {\em arXiv preprint arXiv:2009.14119}, 2020.

\bibitem{bojanowski2016enriching}
Piotr Bojanowski, Edouard Grave, Armand Joulin, and Tomas Mikolov.
\newblock Enriching word vectors with subword information.
\newblock {\em arXiv preprint arXiv:1607.04606}, 2016.

\bibitem{chen2019multi}
Zhao-Min Chen, Xiu-Shen Wei, Xin Jin, and Yanwen Guo.
\newblock Multi-label image recognition with joint class-aware map
  disentangling and label correlation embedding.
\newblock In {\em 2019 IEEE International Conference on Multimedia and Expo
  (ICME)}, pages 622--627. IEEE, 2019.

\bibitem{chen2019multi2}
Zhao-Min Chen, Xiu-Shen Wei, Peng Wang, and Yanwen Guo.
\newblock Multi-label image recognition with graph convolutional networks.
\newblock In {\em Proceedings of the IEEE Conference on Computer Vision and
  Pattern Recognition}, pages 5177--5186, 2019.

\bibitem{nuswide}
Tat-Seng Chua, Jinhui Tang, Richang Hong, Haojie Li, Zhiping Luo, and Yantao
  Zheng.
\newblock Nus-wide: a real-world web image database from national university of
  singapore.
\newblock In {\em CIVR}, 2009.

\bibitem{crammer2001algorithmic}
Koby Crammer and Yoram Singer.
\newblock On the algorithmic implementation of multiclass kernel-based vector
  machines.
\newblock {\em Journal of machine learning research}, 2(Dec):265--292, 2001.

\bibitem{imagenet_cvpr09}
J. Deng, W. Dong, R. Socher, L.-J. Li, K. Li, and L. Fei-Fei.
\newblock {ImageNet: A Large-Scale Hierarchical Image Database}.
\newblock In {\em CVPR09}, 2009.

\bibitem{deutsch2017zero}
Shay Deutsch, Soheil Kolouri, Kyungnam Kim, Yuri Owechko, and Stefano Soatto.
\newblock Zero shot learning via multi-scale manifold regularization.
\newblock In {\em Proceedings of the IEEE conference on computer vision and
  pattern recognition}, pages 7112--7119, 2017.

\bibitem{devries2017improved}
Terrance DeVries and Graham~W. Taylor.
\newblock Improved regularization of convolutional neural networks with cutout,
  2017.

\bibitem{durand2019learning}
Thibaut Durand, Nazanin Mehrasa, and Greg Mori.
\newblock Learning a deep convnet for multi-label classification with partial
  labels.
\newblock In {\em Proceedings of the IEEE Conference on Computer Vision and
  Pattern Recognition}, pages 647--657, 2019.

\bibitem{frome2013devise}
Andrea Frome, Greg~S Corrado, Jon Shlens, Samy Bengio, Jeff Dean, Marc'Aurelio
  Ranzato, and Tomas Mikolov.
\newblock Devise: A deep visual-semantic embedding model.
\newblock In {\em Advances in neural information processing systems}, pages
  2121--2129, 2013.

\bibitem{fu2015transductive}
Yanwei Fu, Yongxin Yang, Tim Hospedales, Tao Xiang, and Shaogang Gong.
\newblock Transductive multi-label zero-shot learning.
\newblock {\em arXiv preprint arXiv:1503.07790}, 2015.

\bibitem{gao2020multi}
Bin-Bin Gao and Hong-Yu Zhou.
\newblock Multi-label image recognition with multi-class attentional regions.
\newblock {\em arXiv preprint arXiv:2007.01755}, 2020.

\bibitem{grave2018learning}
Edouard Grave, Piotr Bojanowski, Prakhar Gupta, Armand Joulin, and Tomas
  Mikolov.
\newblock Learning word vectors for 157 languages.
\newblock In {\em Proceedings of the International Conference on Language
  Resources and Evaluation (LREC 2018)}, 2018.

\bibitem{gupta2021generative}
Akshita Gupta, Sanath Narayan, Salman Khan, Fahad~Shahbaz Khan, Ling Shao, and
  Joost van~de Weijer.
\newblock Generative multi-label zero-shot learning.
\newblock {\em arXiv preprint arXiv:2101.11606}, 2021.

\bibitem{hayat2020synthesizing}
Nasir Hayat, Munawar Hayat, Shafin Rahman, Salman Khan, Syed~Waqas Zamir, and
  Fahad~Shahbaz Khan.
\newblock Synthesizing the unseen for zero-shot object detection.
\newblock In {\em ACCV}, 2020.

\bibitem{he2016deep}
Kaiming He, Xiangyu Zhang, Shaoqing Ren, and Jian Sun.
\newblock Deep residual learning for image recognition.
\newblock In {\em Proceedings of the IEEE conference on computer vision and
  pattern recognition}, pages 770--778, 2016.

\bibitem{huynh2020shared}
Dat Huynh and Ehsan Elhamifar.
\newblock A shared multi-attention framework for multi-label zero-shot
  learning.
\newblock In {\em Proceedings of the IEEE/CVF Conference on Computer Vision and
  Pattern Recognition}, pages 8776--8786, 2020.

\bibitem{jayaraman2014zero}
Dinesh Jayaraman and Kristen Grauman.
\newblock Zero-shot recognition with unreliable attributes.
\newblock {\em Advances in neural information processing systems},
  27:3464--3472, 2014.

\bibitem{kim2018bilinear}
Jin-Hwa Kim, Jaehyun Jun, and Byoung-Tak Zhang.
\newblock Bilinear attention networks.
\newblock In {\em NeurIPS}, 2018.

\bibitem{kingma2017adam}
Diederik~P. Kingma and Jimmy Ba.
\newblock Adam: A method for stochastic optimization, 2017.

\bibitem{kodirov2017semantic}
Elyor Kodirov, Tao Xiang, and Shaogang Gong.
\newblock Semantic autoencoder for zero-shot learning.
\newblock In {\em Proceedings of the IEEE Conference on Computer Vision and
  Pattern Recognition}, pages 3174--3183, 2017.

\bibitem{openimages}
Alina Kuznetsova, Hassan Rom, Neil Alldrin, Jasper Uijlings, Ivan Krasin, Jordi
  Pont-Tuset, Shahab Kamali, Stefan Popov, Matteo Malloci, Tom Duerig, et~al.
\newblock The open images dataset v4: Unified image classification, object
  detection, and visual relationship detection at scale.
\newblock {\em arXiv preprint arXiv:1811.00982}, 2018.

\bibitem{lee2018multi}
Chung-Wei Lee, Wei Fang, Chih-Kuan Yeh, and Yu-Chiang Frank~Wang.
\newblock Multi-label zero-shot learning with structured knowledge graphs.
\newblock In {\em Proceedings of the IEEE conference on computer vision and
  pattern recognition}, pages 1576--1585, 2018.

\bibitem{li2015zero}
Xirong Li, Shuai Liao, Weiyu Lan, Xiaoyong Du, and Gang Yang.
\newblock Zero-shot image tagging by hierarchical semantic embedding.
\newblock In {\em Proceedings of the 38th International ACM SIGIR Conference on
  Research and Development in Information Retrieval}, pages 879--882, 2015.

\bibitem{li2017zero}
Yanan Li, Donghui Wang, Huanhang Hu, Yuetan Lin, and Yueting Zhuang.
\newblock Zero-shot recognition using dual visual-semantic mapping paths.
\newblock In {\em Proceedings of the IEEE Conference on Computer Vision and
  Pattern Recognition}, pages 3279--3287, 2017.

\bibitem{coco}
Tsung-Yi Lin, Michael Maire, Serge Belongie, James Hays, Pietro Perona, Deva
  Ramanan, Piotr Doll{\'a}r, and C~Lawrence Zitnick.
\newblock Microsoft coco: Common objects in context.
\newblock In {\em ECCV}, 2014.

\bibitem{loshchilov2019decoupled}
Ilya Loshchilov and Frank Hutter.
\newblock Decoupled weight decay regularization, 2019.

\bibitem{mikolov2013efficient}
Tomas Mikolov, Kai Chen, Greg Corrado, and Jeffrey Dean.
\newblock Efficient estimation of word representations in vector space.
\newblock {\em arXiv preprint arXiv:1301.3781}, 2013.

\bibitem{miller1995wordnet}
George~A Miller.
\newblock Wordnet: a lexical database for english.
\newblock {\em Communications of the ACM}, 38(11):39--41, 1995.

\bibitem{morgado2017semantically}
Pedro Morgado and Nuno Vasconcelos.
\newblock Semantically consistent regularization for zero-shot recognition.
\newblock In {\em Proceedings of the IEEE Conference on Computer Vision and
  Pattern Recognition}, pages 6060--6069, 2017.

\bibitem{norouzi2013zero}
Mohammad Norouzi, Tomas Mikolov, Samy Bengio, Yoram Singer, Jonathon Shlens,
  Andrea Frome, Greg~S Corrado, and Jeffrey Dean.
\newblock Zero-shot learning by convex combination of semantic embeddings.
\newblock {\em arXiv preprint arXiv:1312.5650}, 2013.

\bibitem{pennington2014glove}
Jeffrey Pennington, Richard Socher, and Christopher~D Manning.
\newblock Glove: Global vectors for word representation.
\newblock In {\em Proceedings of the 2014 conference on empirical methods in
  natural language processing (EMNLP)}, pages 1532--1543, 2014.

\bibitem{rahman2019deep0tag}
Shafin Rahman, Salman Khan, and Nick Barnes.
\newblock Deep0tag: Deep multiple instance learning for zero-shot image
  tagging.
\newblock {\em IEEE Transactions on Multimedia}, 22(1):242--255, 2019.

\bibitem{rahman2019transductive}
Shafin Rahman, Salman Khan, and Nick Barnes.
\newblock Transductive learning for zero-shot object detection.
\newblock In {\em Proceedings of the IEEE International Conference on Computer
  Vision}, pages 6082--6091, 2019.

\bibitem{rahman2018zero}
Shafin Rahman, Salman Khan, and Fatih Porikli.
\newblock Zero-shot object detection: Learning to simultaneously recognize and
  localize novel concepts.
\newblock In {\em Asian Conference on Computer Vision}, pages 547--563.
  Springer, 2018.

\bibitem{ren2017multiple}
Zhou Ren, Hailin Jin, Zhe Lin, Chen Fang, and Alan~L Yuille.
\newblock Multiple instance visual-semantic embedding.
\newblock In {\em BMVC}, 2017.

\bibitem{ridnik2021tresnet}
Tal Ridnik, Hussam Lawen, Asaf Noy, Emanuel Ben~Baruch, Gilad Sharir, and
  Itamar Friedman.
\newblock Tresnet: High performance gpu-dedicated architecture.
\newblock In {\em Proceedings of the IEEE/CVF Winter Conference on Applications
  of Computer Vision}, pages 1400--1409, 2021.

\bibitem{romera2015embarrassingly}
Bernardino Romera-Paredes and Philip Torr.
\newblock An embarrassingly simple approach to zero-shot learning.
\newblock In {\em International Conference on Machine Learning}, pages
  2152--2161, 2015.

\bibitem{schonfeld2019generalized}
Edgar Schonfeld, Sayna Ebrahimi, Samarth Sinha, Trevor Darrell, and Zeynep
  Akata.
\newblock Generalized zero-and few-shot learning via aligned variational
  autoencoders.
\newblock In {\em Proceedings of the IEEE Conference on Computer Vision and
  Pattern Recognition}, pages 8247--8255, 2019.

\bibitem{simonyan2014very}
Karen Simonyan and Andrew Zisserman.
\newblock Very deep convolutional networks for large-scale image recognition.
\newblock {\em arXiv preprint arXiv:1409.1556}, 2014.

\bibitem{smith2018disciplined}
Leslie~N. Smith.
\newblock A disciplined approach to neural network hyper-parameters: Part 1 --
  learning rate, batch size, momentum, and weight decay, 2018.

\bibitem{wang2020multi}
Ya Wang, Dongliang He, Fu Li, Xiang Long, Zhichao Zhou, Jinwen Ma, and Shilei
  Wen.
\newblock Multi-label classification with label graph superimposing.
\newblock In {\em Proceedings of the AAAI Conference on Artificial
  Intelligence}, volume~34, pages 12265--12272, 2020.

\bibitem{wang2017multi}
Zhouxia Wang, Tianshui Chen, Guanbin Li, Ruijia Xu, and Liang Lin.
\newblock Multi-label image recognition by recurrently discovering attentional
  regions.
\newblock In {\em Proceedings of the IEEE international conference on computer
  vision}, pages 464--472, 2017.

\bibitem{xian2016latent}
Yongqin Xian, Zeynep Akata, Gaurav Sharma, Quynh Nguyen, Matthias Hein, and
  Bernt Schiele.
\newblock Latent embeddings for zero-shot classification.
\newblock In {\em Proceedings of the IEEE Conference on Computer Vision and
  Pattern Recognition}, pages 69--77, 2016.

\bibitem{xian2018feature}
Yongqin Xian, Tobias Lorenz, Bernt Schiele, and Zeynep Akata.
\newblock Feature generating networks for zero-shot learning.
\newblock In {\em Proceedings of the IEEE conference on computer vision and
  pattern recognition}, pages 5542--5551, 2018.

\bibitem{xian2017zero}
Yongqin Xian, Bernt Schiele, and Zeynep Akata.
\newblock Zero-shot learning-the good, the bad and the ugly.
\newblock In {\em Proceedings of the IEEE Conference on Computer Vision and
  Pattern Recognition}, pages 4582--4591, 2017.

\bibitem{ye2020attention}
Jin Ye, Junjun He, Xiaojiang Peng, Wenhao Wu, and Yu Qiao.
\newblock Attention-driven dynamic graph convolutional network for multi-label
  image recognition.
\newblock In {\em European Conference on Computer Vision}, pages 649--665.
  Springer, 2020.

\bibitem{yeh2019multilabel}
Mei-Chen Yeh and Yi-Nan Li.
\newblock Multilabel deep visual-semantic embedding.
\newblock {\em IEEE transactions on pattern analysis and machine intelligence},
  42(6):1530--1536, 2019.

\bibitem{you2020cross}
Renchun You, Zhiyao Guo, Lei Cui, Xiang Long, Yingze Bao, and Shilei Wen.
\newblock Cross-modality attention with semantic graph embedding for
  multi-label classification.
\newblock In {\em AAAI}, pages 12709--12716, 2020.

\bibitem{zhang2017learning}
Li Zhang, Tao Xiang, and Shaogang Gong.
\newblock Learning a deep embedding model for zero-shot learning.
\newblock In {\em Proceedings of the IEEE Conference on Computer Vision and
  Pattern Recognition}, pages 2021--2030, 2017.

\bibitem{zhang2016fast}
Yang Zhang, Boqing Gong, and Mubarak Shah.
\newblock Fast zero-shot image tagging.
\newblock In {\em 2016 IEEE Conference on Computer Vision and Pattern
  Recognition (CVPR)}, pages 5985--5994. IEEE, 2016.

\end{thebibliography}
}

\newpage
\clearpage
\appendix
\newpage

\begin{center} \begin{Large}{Appendix} \end{Large} \end{center}
\label{appendix}

\section{More Experiments}
\label{sec:more_experiments}
\subsection{Backbone Variations}

In our experiments we use TResNet-M \cite{ridnik2021tresnet} as a backbone for our visual model, due to its efficiency and reported high accuracy on several competitive computer vision datasets. To further extend our analysis and comparison with prior works we also explore two popular backbone architectures, VGG19 \cite{simonyan2014very} and ResNet50 \cite{he2016deep} in Table \ref{tab:backbone}. We report results using our approach as well as adding a comparison to Fast0Tag \cite{zhang2016fast} loss function with our E2E training scheme as a baseline. As can be seen, using our approach with VGG19 as a backbone, the results in terms of mAP for both zero-shot and generalized zero-shot are superior compared to prior works but lower than our current backbone, while using ResNet50 as a backbone improves over VGG19 in all metrics. Best results are achieved using TResNet-M backbone. In addition it can also be seen that the results in terms of mAP for tag-based image retrieval using different backbone variations are higher than current prior works, suggesting that our training scheme extends and may improve the quality of various model architectures. 

\begin{table}[h!]
\centering
\caption{Results using alternative backbones on NUS-WIDE test set. We report the results in terms of F1($K=3$), F1($K=5$), and mAP for ZSL and GZSL tasks. Best results are in bold.}\vspace{0.2em} 
\setlength{\tabcolsep}{12pt}
\adjustbox{width=1\linewidth}{
\begin{tabular}{cccccc} 
\toprule[0.15em]
\textbf{Backbone} & \textbf{Method} &\textbf{Task} & \begin{tabular}[c]{@{}c@{}} \textbf{F1($K=3$)} \end{tabular} & \begin{tabular}[c]{@{}c@{}} \textbf{F1($K=5$)} \end{tabular} & \textbf{mAP} \\
\toprule[0.15em]
\multirow{2}{*}{VGG19 \cite{simonyan2014very}} & \multirow{2}{*}{Fast0Tag \cite{zhang2016fast}} & ZSL & 24.2 & 22.2 & 20.2 \\
 & &GZSL & 11.7 & 13.0 & 6.6 \\ 
\cmidrule(r){3-6}
\multirow{2}{*}{TResNet-M \cite{ridnik2021tresnet}} & \multirow{2}{*}{Fast0Tag \cite{zhang2016fast}} & ZSL & 25.7 & 23.3 & 21.6 \\
 & &GZSL & 15.4 & 16.6 & 9.7 \\ 
\cmidrule(r){3-6}
\multirow{2}{*}{VGG19 \cite{simonyan2014very}} & \multirow{2}{*}{Ours} & ZSL & 29.0 & 26.5 & 24.2 \\
 & &GZSL & 16.8 & 19.0 & 9.9 \\ 
\cmidrule(r){3-6}
\multirow{2}{*}{ResNet50 \cite{he2016deep}} & \multirow{2}{*}{Ours} &ZSL & 30.0 & 27.6 & 24.4 \\
 & &GZSL & 17.7 & 20.1 & 11.2 \\ 
\cmidrule(r){3-6}
\multirow{2}{*}{{TResNet-M \cite{ridnik2021tresnet}}} & \multirow{2}{*}{Ours} &ZSL & \textbf{{30.5}} & \textbf{27.8} & \textbf{{25.9}} \\
 & &GZSL & {\textbf{18.5}} & \textbf{{21.0}} & \textbf{{12.1}} \\
 
\bottomrule[0.1em]
\end{tabular}%
}
\vspace{-0.2cm}
\label{tab:backbone}
\end{table}

\section{Reproduciblity}
To support future research in the field, we currently work to publish our trained models and share a fully reproducible training code on GitHub.

\section{Additional Qualitative Results}
\begin{figure*}
\centering
\begin{tabular}{lll}
\subcaptionbox*{\textbf{graffiti} \\ \textbf{art} \\ \textbf{London} \\ mural \\ England \\ urban \\ green \\ paint \\ war \\ politics}{\includegraphics[height = 2.6cm]{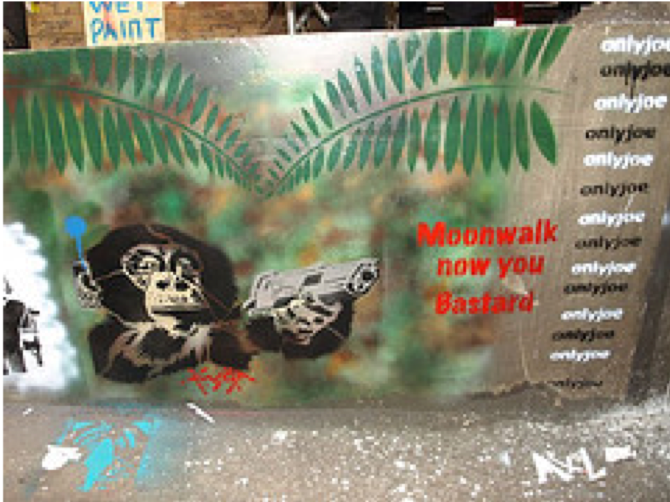}} &
\subcaptionbox*{officers \\ protesters \\ riot \\ politics \\ *\textbf{police}* \\ \textbf{London} \\ men \\ roadblock \\ *protest* \\ soldier}{\includegraphics[height = 2.6cm]{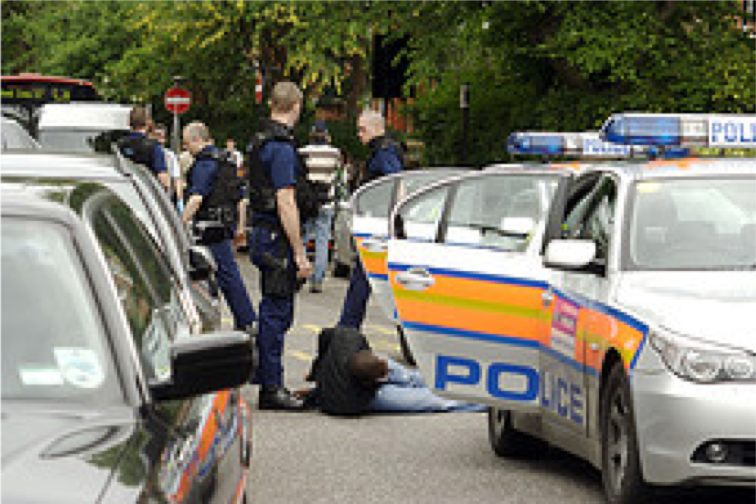}} &
\subcaptionbox*{firefighter \\ \textbf{demonstration} \\ France \\ Canada \\ \textbf{riot} \\ action \\ officers \\ winter \\ sport \\ *\textbf{protest}* \\ }{\includegraphics[height = 2.6cm]{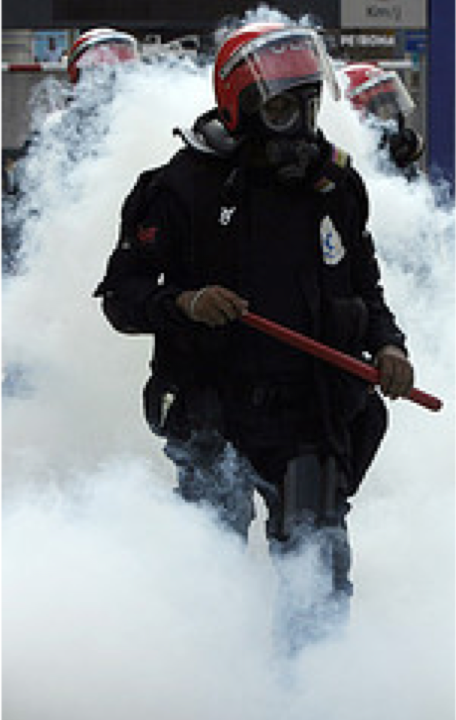}} \\
\subcaptionbox*{wildlife \\ nature \\ deer \\ \textbf{moose} \\ *elk* \\ Canada \\ wild \\ animals \\ park \\ \textbf{Alaska}}{\includegraphics[height = 2.6cm]{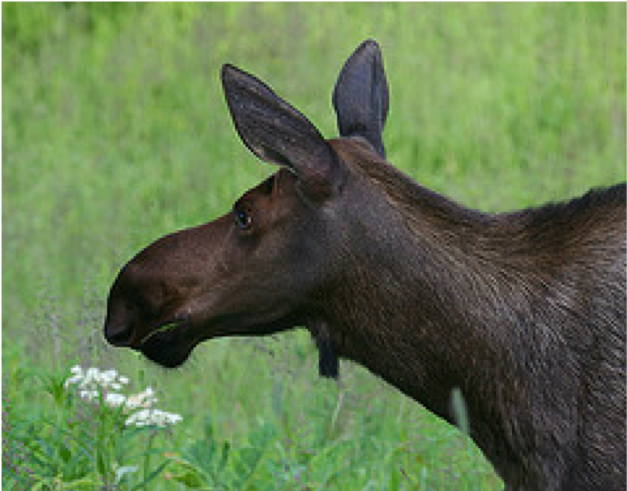}} &

\subcaptionbox*{football \\ crowd \\ \textbf{cheering} \\ *soccer* \\ baseball \\ red \\ England \\ game \\ parade \\ basketball}{\includegraphics[height = 2.6cm]{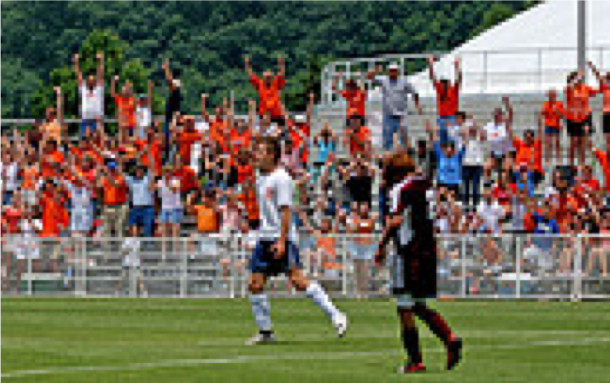}} &

\subcaptionbox*{\textbf{bride} \\ *\textbf{wedding}* \\ Hawaii \\ sea \\ \textbf{bravo} \\ \textbf{beautiful} \\ white \\ couple \\ groom \\ dress }{\includegraphics[height = 2.6cm]{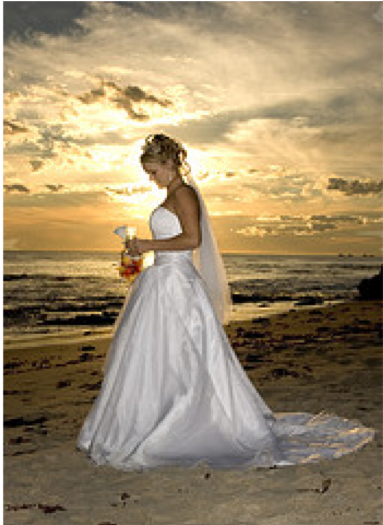}}

\end{tabular}
\caption{Qualitative results showing the top-10 tags retrieved using our proposed method. Bold text represents the correct tags according to the provided ground truth in NUS-WIDE test set. Asterisks mark unseen tags.}
\label{fig:more_qualitative}
\end{figure*}

We present in figure \ref{fig:more_qualitative} additional qualitative results using our proposed method for several sample images from NUS-WIDE test set. It can be seen that in several cases the unseen tags (marked by asterisks) are ranked in the top-10. In addition, while some of the unseen tags are incorrect based on the ground truth annotation, in most cases there exists a noticeable semantic relation between these tags to the image.

\end{document}